\documentclass[11pt]{article}
\usepackage[preprint]{acl}
\usepackage{times,latexsym,amsmath,amssymb,booktabs,graphicx}
\usepackage[T1]{fontenc}
\usepackage[utf8]{inputenc}
\usepackage{microtype,colortbl}
\title{Faster and Better Query (FBQ)\\Learning Future-Position Probes from the First Generated Token}
\author{Jiankun Zou \qquad Daixing Chen \qquad Tongtong Yuan\textsuperscript{*}\\
  \normalfont Beijing University of Technology\\
  {\small\href{mailto:zoujiankun2006@emails.bjut.edu.cn}{\texttt{zoujiankun2006@emails.bjut.edu.cn}}\quad
  \href{mailto:chendaixing@emails.bjut.edu.cn}{\texttt{chendaixing@emails.bjut.edu.cn}}}\\
  {\small\textsuperscript{*}Corresponding author\quad\href{mailto:yuantt@bjut.edu.cn}{\texttt{yuantt@bjut.edu.cn}}}}
\hypersetup{
  pdfauthor={Jiankun Zou, Daixing Chen, Tongtong Yuan},
  pdftitle={Faster and Better Query (FBQ): Learning Future-Position Probes from the First Generated Token}}
\newcommand{\fbq}{\textsc{FBQ}}
\begin{document}
\maketitle

\begin{abstract}
When a large language model (LLM) evicts prompt keys and values at the end of prefill, it must decide what to keep before it sees the queries from subsequent decoding steps. Generating a draft continuation can approximate those queries at the cost of extra autoregressive steps. Can we use the embedding of the first output token, already available from the final prefill logits, to construct low-cost pseudo queries that approximate future decoding queries? We start with FTQ-8, which places eight copies of the first token ID at consecutive future positions and reads their attention to the prompt in one causal forward pass. Each copy sees a different prefix, but a generic opening such as \emph{the} offers a poor literal guess for seven successive tokens. Faster and Better Query (\fbq{}) preserves the first embedding and learns an additive offset at each of the seven remaining positions. We train these offsets to predict the average prompt attention of a full-cache continuation. On LongBench, FBQ performs comparably to eight-step Lookahead Q-Cache at the same cache budget, using a single forward pass to compute all probe queries.
\end{abstract}

\let\usualTopFraction\topfraction
\renewcommand{\topfraction}{.85}
\AddToHookNext{shipout/after}{\global\let\topfraction\usualTopFraction}
\suppressfloats[t]
\begin{figure}[t]
\centering
\includegraphics[width=\columnwidth]{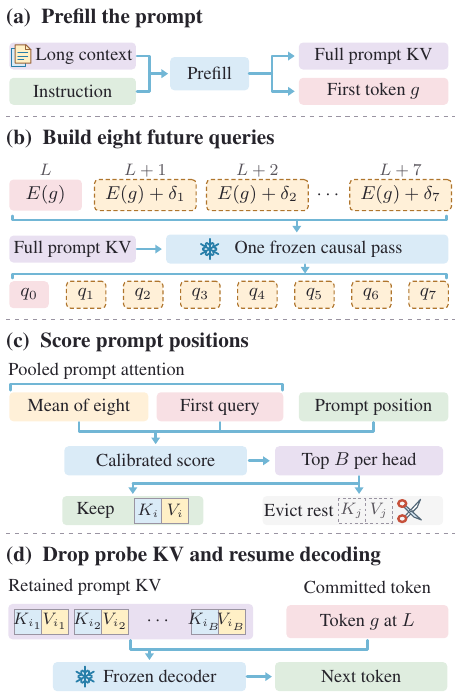}
\caption{FBQ uses the first output token $g$ to select a prompt cache. FTQ-8 repeats $E(g)$ at all eight positions. FBQ learns seven offsets offline from full-cache continuation attention. After selection, it discards the probe KV and processes $g$ at its original position $L$.}
\label{fig:overview}
\end{figure}

\section{Introduction}
Consider two questions about the same long report: one asks who approved an expenditure, the other asks for the report's main findings. A model might begin both answers with \emph{the}. A cache eviction rule must decide, before either answer unfolds, which parts of the report will still be available when the model generates a name or summarizes a later section. The first output token is a particularly interesting signal at this point. Its identity reflects a decision the model has already made using the full prompt. Its attention query, once computed in the context of the question, can differ across the two requests even when their first token IDs coincide.

Lookahead Q-Cache (LAQ) estimates the importance of prompt positions from pseudo lookahead queries \citep{wang2025}. Generating a draft continuation gives the queries plausible lexical context, at the cost of a sequential dependency between draft steps. We explore how far the first committed output token alone can take this idea. Our starting point, FTQ-8, inserts its ID into eight consecutive positions after the prompt on a temporary cache branch. All eight inputs are available immediately. With a causal mask, each later copy can attend to a longer prefix of the block, and the model applies the positional encoding for its actual future position. Thus repeating an ID yields eight contextualized queries, even though the method predicts only one vocabulary item.

Repetition also imposes a strong lexical assumption. A model that starts an answer with \emph{this} may next produce a noun, a verb, or a reference to a passage. FTQ-8 supplies the embedding of \emph{this} in each case. We retain the part of FTQ-8 that is grounded in the model's own decision and relax the part that extrapolates it. In Faster and Better Query (\fbq{}, Figure~\ref{fig:overview}), the first position takes the committed token unchanged. Each later position starts from the same embedding and adds a learned vector. These offsets can change which prompt evidence a probe attends to without requiring the probe to decode a plausible word.

An exploratory run first motivated this construction. Across 16 tasks, eight repeated positions scored 35.06, compared with 34.69 for 32 repeated positions in the same result series. A residual-probe variant in that series scored 35.37. These observations favored investigating the contents of a short probe block over simply lengthening it. We report that pilot separately from our main LAQ comparison because its full-cache reference was 38.72. In the main experiment, the reference is 48.07. All comparisons below specify the result series from which the numbers come.

\section{From a token ID to future queries}
\paragraph{FTQ-8.}
Let $x$ be a prompt of length $L$ and let $g$ be the first output token selected from its final prefill logits under the decoding policy. On a branch of the prompt cache, place $g$ at positions $L,\ldots,L+7$ (zero-based). With input embedding table $E$, all eight rows are $E(g)$. For row $t$, the causal mask exposes the prompt and rows $0,\ldots,t$. At row zero the frozen model therefore computes the query that processing the real first output token would produce on the full prompt. For later rows, the state depends on both the new position and the representations of earlier copies. They probe subsequent decoding positions in one pass. The generated answer receives only the committed token $g$.

\paragraph{FBQ.}
For embedding dimension $d$, introduce seven vectors $\delta_1,\ldots,\delta_7\in\mathbb{R}^d$. The temporary input block is
\begin{align}
 u_0 &= E(g), \nonumber\\
 u_t &= E(g)+\delta_t,\qquad 1\leq t\leq 7. \label{eq:probe}
\end{align}
The vectors are shared across requests and initialized to zero, which recovers the FTQ-8 inputs exactly. The base embedding changes with $g$ and every row still attends to the particular prompt. For two requests with the same opening token, the input blocks coincide but their queries can differ. The causal mask also prevents any learned offset from altering the first query. This gives the scorer one unchanged observation of the actual decoding trajectory throughout training.

\begin{figure*}[t]
\centering
\begin{minipage}[c]{.492\textwidth}\centering
\includegraphics[width=\linewidth]{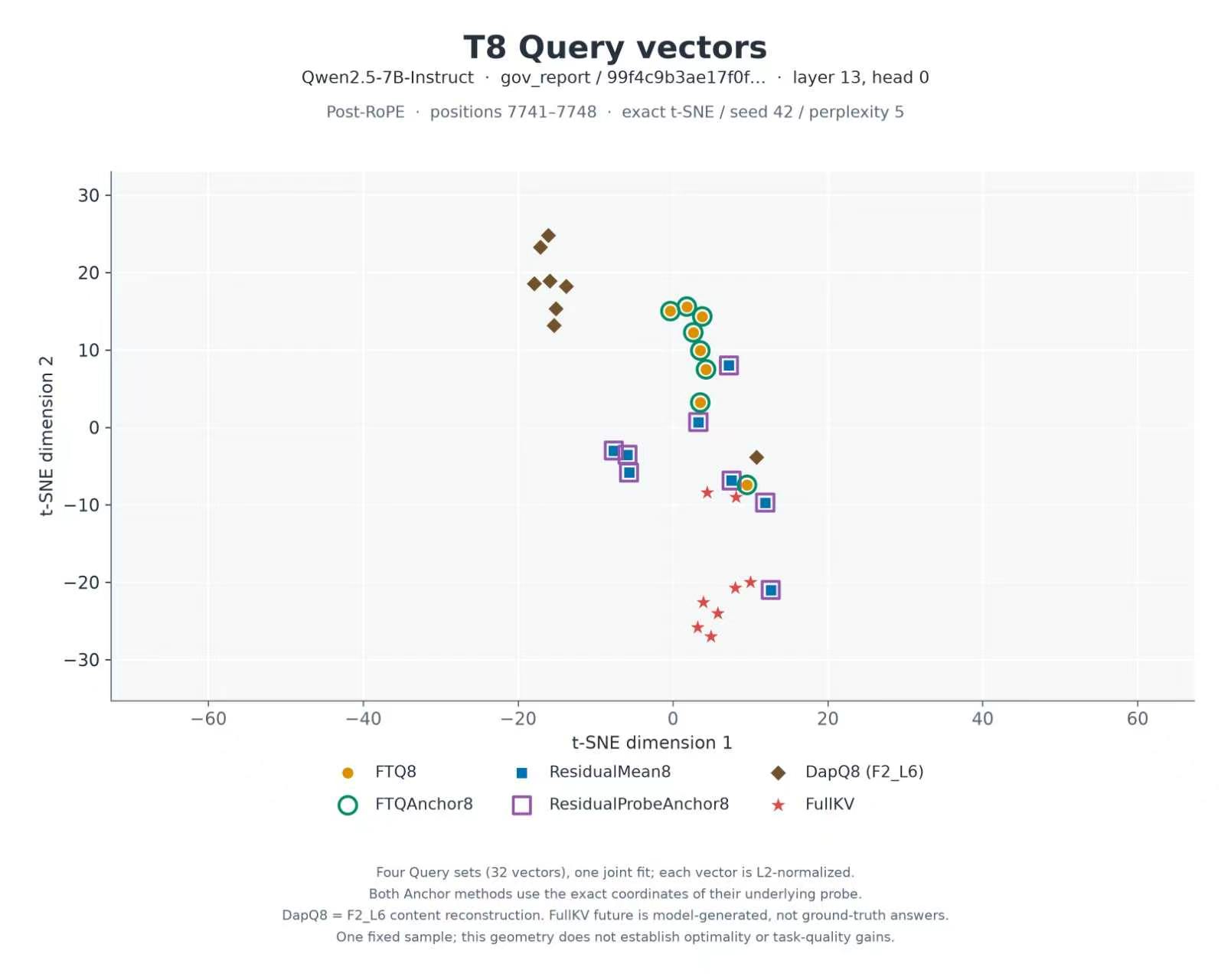}
\end{minipage}\hfill
\begin{minipage}[c]{.492\textwidth}\centering
\includegraphics[width=\linewidth]{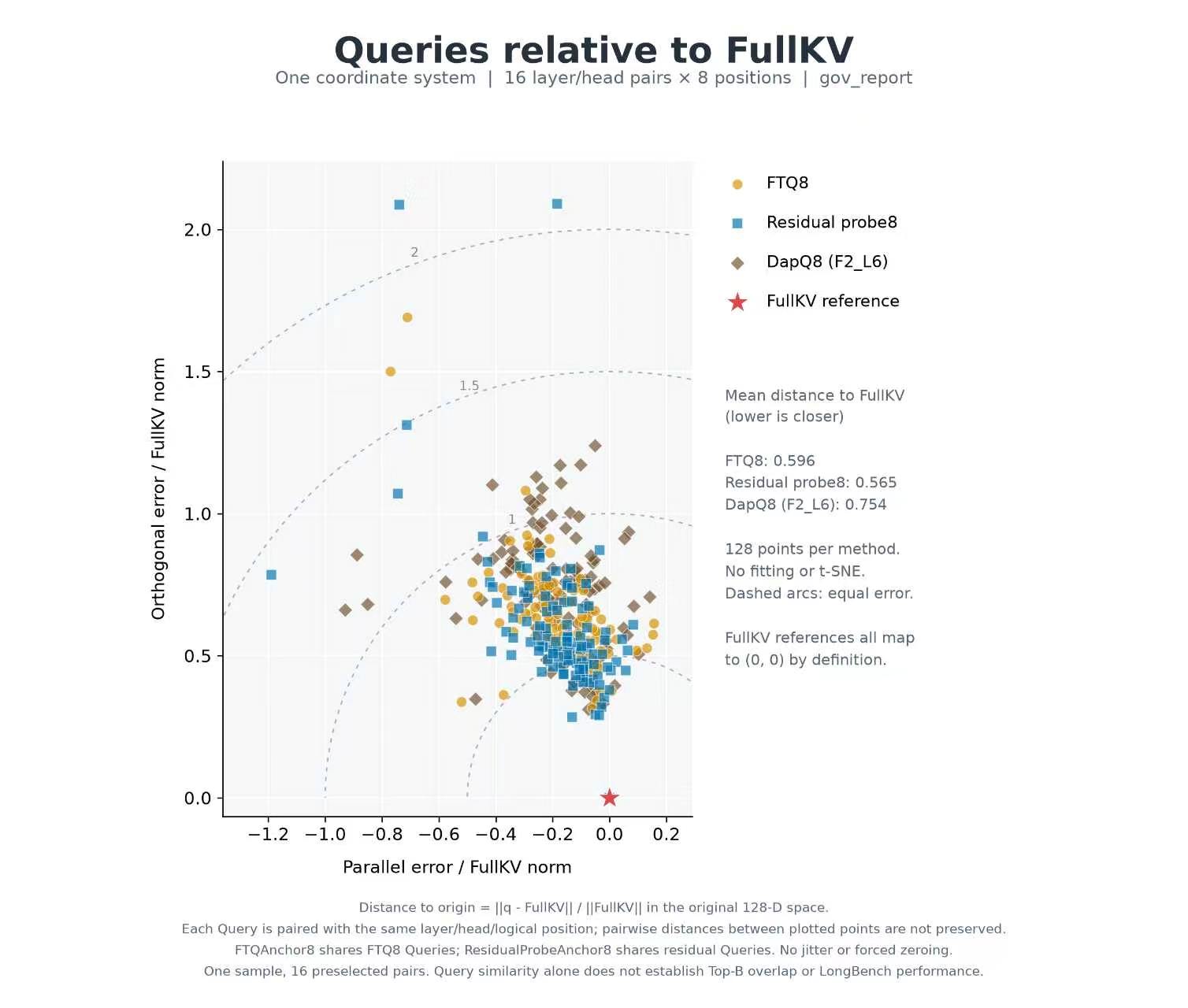}
\end{minipage}
\caption{Query diagnostics on a GovReport example with Qwen2.5-7B-Instruct. Left: a joint t-SNE of post-RoPE queries at eight future positions for layer 13, head 0. Matching outline markers reuse the queries before reranking. Right: parallel and orthogonal errors relative to the corresponding FullKV query, normalized by its norm, for 16 layer/head pairs at eight positions each. Each point's radius equals its original relative error.}
\label{fig:queries}
\end{figure*}

\paragraph{A query-level observation.}
Figure~\ref{fig:queries} compares queries with those from a continuation generated using the full prompt cache. The repeated-ID construction yields distinct queries at successive positions in the head shown. To quantify alignment, we pair each probe query with the full-cache query at the same layer, head and position and measure relative Euclidean distance in the original 128-dimensional space. Across 128 such pairs on this example, the mean is 0.596 for FTQ-8, 0.565 for the residual probes and 0.754 for the DapQ8 content-reconstruction control. The residual reduces the mean distance by 0.031, about 5.2\% relative to FTQ-8. These are measurements within one prompt. The 128 pairs share that prompt, and their distances measure query alignment rather than task accuracy.

\paragraph{What the offsets learn.}
Offline, the full-cache model generates a continuation. Its first 32 decoding positions provide a teacher map of attention to the prompt. For a fixed layer and head, write $a_t(i)$ for a probe's attention to prompt position $i$. The softmax includes prompt keys and causally visible probe keys. We take the prompt slice of the resulting probabilities and apply a width-five average pool along the prompt after averaging over the eight rows:
\begin{equation}
 s_\delta(i)=\operatorname{Pool}_5\!\left[
          \tfrac{1}{8}\textstyle\sum_{t=0}^7 a_t
          \right](i).
\end{equation}
We construct $s^\star$ analogously using the teacher's 32 genuine continuation positions. For each document, we average $(s_\delta(i)-s^\star(i))^2$ over prompt positions, heads and layers, then average over documents. Gradients pass through the frozen transformer to the seven offsets. The teacher supplies attention targets offline. At inference, the response prefix required to construct the probes consists solely of $g$.

\paragraph{Selecting and using the cache.}
After fitting the offsets, we freeze them and calibrate a linear scorer against membership in the teacher's top-$B$ prompt positions. The scorer uses both the pooled mean $s_\delta$ and the pooled attention of row zero. Each map supplies a rank percentile and a log score relative to its head mean. Normalized prompt position and a bias give six features in total. This combination lets the scorer preserve evidence favored by the first genuine query even when averaging the block lowers its rank. We keep the $B$ highest-scoring prompt positions for each layer and query head, restore their original order, discard the temporary branch, and continue decoding from the retained prompt cache with the committed first token. Generated tokens subsequently add their own cache entries.

\paragraph{Cost of the probe.}
For a prompt of length $L$, one head evaluates $8L$ prompt-query dot products for the probe block, as well as attention within the short causal suffix. Batching changes the sequential dependence of these computations. All eight embeddings are specified by Eq.~\ref{eq:probe} before the first layer runs. In a draft that chooses each subsequent token from the previous step's logits, those token choices serialize the corresponding forwards. FBQ still incurs a post-prefill forward and reads the prompt cache to score it. Wall-clock savings depend on how the implementation handles these reads and the temporary branch.

\section{Experimental results}
We evaluate on the 16-task English and code subset of LongBench \citep{bai2024}. It covers single- and multi-document question answering, summarization, few-shot learning, synthetic tasks and code completion. The main comparison uses Mistral-7B v0.3 and a history budget of 128 prompt positions per head for the compressed methods. Each method covers 3,750 examples. We report the arithmetic mean of the 16 task scores, giving each task equal weight. FullKV retains the entire prompt. LAQ-2 and LAQ-8 use two and eight lookahead steps. We show all three evaluated residual checkpoints so that the spread across training epochs is visible.

\begin{table*}[t]
\centering
\begingroup
\setlength{\tabcolsep}{1.65pt}
\renewcommand{\arraystretch}{1.13}
\newcommand{\taskname}[1]{\rotatebox{53}{\strut\bfseries #1}}
\newcommand{\band}[1]{\rowcolor[gray]{0.93}\multicolumn{18}{c}{\strut\bfseries #1}}
\fontsize{7.3}{8.4}\selectfont
\resizebox{\textwidth}{!}{%
\begin{tabular}{@{}l*{16}{r}r@{}}
\toprule
& \multicolumn{3}{c}{\bfseries Single-Document QA} & \multicolumn{3}{c}{\bfseries Multi-Document QA} & \multicolumn{3}{c}{\bfseries Summarization} & \multicolumn{3}{c}{\bfseries Few-shot Learning} & \multicolumn{2}{c}{\bfseries Synthetic} & \multicolumn{2}{c}{\bfseries Code} & \\
\cmidrule(lr){2-4}\cmidrule(lr){5-7}\cmidrule(lr){8-10}\cmidrule(lr){11-13}\cmidrule(lr){14-15}\cmidrule(lr){16-17}
\textbf{Method} & \taskname{NrtvQA} & \taskname{Qasper} & \taskname{MF-en} & \taskname{HotpotQA} & \taskname{2WikiMQA} & \taskname{MuSiQue} & \taskname{GovReport} & \taskname{QMSum} & \taskname{MultiNews} & \taskname{TREC} & \taskname{TriviaQA} & \taskname{SAMSum} & \taskname{PCount} & \taskname{PRe} & \taskname{LCC} & \taskname{RB-P} & \textbf{Avg.}\\
\midrule
\band{KV cache size = 128: published results (Judge Q, Table 2)}\\
FullKV & 29.20 & 41.58 & 53.11 & 49.27 & 39.01 & 28.58 & 31.95 & 25.67 & 23.89 & 76.00 & 88.59 & 47.50 & 5.50 & 98.00 & 51.97 & 46.28 & 46.01\\
\cmidrule{1-18}
StreamingLLM & 21.42 & 20.42 & 30.56 & 40.83 & 32.77 & 18.26 & 18.12 & 19.70 & 16.42 & 46.00 & 81.07 & 31.78 & 6.00 & 81.00 & 37.61 & 34.88 & 33.55\\
H$_2$O & 24.48 & 28.97 & 43.61 & 42.98 & 35.31 & 24.03 & 17.46 & 23.04 & 17.60 & 38.50 & 87.67 & 43.57 & \textbf{6.50} & \textbf{95.50} & 47.60 & 41.52 & 38.65\\
SnapKV & 23.86 & 28.02 & 51.44 & 47.71 & 35.90 & 24.16 & 16.53 & 21.79 & 16.70 & 47.50 & 89.54 & 42.98 & 6.00 & 93.50 & 48.87 & \textbf{41.56} & 39.75\\
PyramidKV & 24.21 & 27.04 & 49.20 & 47.52 & 34.81 & 23.51 & 16.78 & 21.39 & 16.87 & 48.00 & 88.16 & 43.05 & 5.50 & 94.00 & \textbf{49.08} & 40.93 & 39.38\\
Judge Q & \textbf{27.86} & \textbf{35.71} & \textbf{54.33} & \textbf{48.14} & \textbf{38.69} & \textbf{26.97} & \textbf{20.43} & \textbf{24.23} & \textbf{19.96} & \textbf{56.50} & \textbf{90.16} & \textbf{46.29} & 5.50 & 93.00 & 48.74 & 41.55 & \textbf{42.38}\\
\band{KV cache size = 128: our evaluation}\\
FullKV & 29.06 & 41.58 & 52.88 & 49.37 & 39.51 & 28.58 & 34.97 & 25.64 & 27.80 & 76.00 & 88.59 & 47.49 & 5.00 & 98.50 & 61.52 & 62.56 & 48.07\\
\cmidrule{1-18}
LAQ-2 & 29.86 & 35.32 & 51.15 & 48.78 & 38.11 & 28.26 & 22.89 & 22.83 & 22.39 & 75.00 & 89.22 & 43.97 & 5.50 & \textbf{95.50} & 55.37 & 53.97 & 44.88\\
LAQ-8 & \textbf{30.06} & \textbf{37.04} & 52.46 & 48.75 & 38.66 & \textbf{29.18} & 22.75 & \textbf{23.96} & 23.06 & 75.00 & 89.72 & 45.38 & 4.50 & \textbf{95.50} & \textbf{57.91} & 56.60 & 45.66\\
FBQ (e2) & 28.75 & 35.85 & 52.39 & 49.26 & 38.40 & 28.45 & 23.34 & 23.77 & 23.31 & \textbf{76.00} & \textbf{89.96} & 45.49 & \textbf{6.00} & 94.00 & 56.68 & 56.49 & 45.51\\
FBQ (e3) & 29.03 & 36.17 & 52.65 & \textbf{49.27} & 38.90 & 28.69 & \textbf{23.56} & 23.84 & \textbf{23.73} & \textbf{76.00} & 89.46 & 45.61 & \textbf{6.00} & 95.00 & 56.83 & \textbf{56.99} & \textbf{45.73}\\
FBQ (e4) & 29.11 & 36.11 & \textbf{53.33} & 49.20 & \textbf{39.00} & 28.34 & 23.43 & 23.75 & 23.10 & \textbf{76.00} & \textbf{89.96} & \textbf{45.67} & \textbf{6.00} & \textbf{95.50} & 56.65 & 56.50 & \textbf{45.73}\\
\bottomrule
\end{tabular}}
\endgroup
\caption{LongBench results on Mistral-7B v0.3, grouped by task family. The upper block reproduces the Mistral-7B-Instruct-v0.3 results of \citet{liu2025} (Table~2). The lower block gives our runs. Each FullKV row is the uncompressed reference for its block. Bold marks the highest displayed score among compressed methods within each block, including ties. Avg.\ is the 16-task mean reported by the corresponding source.}
\label{tab:main}
\end{table*}

\paragraph{Within-protocol comparison.}
FBQ reaches 45.73 at epochs three and four, 0.85 above LAQ-2 and 0.07 above LAQ-8 at the reported precision (Table~\ref{tab:main}). Epoch two is 0.15 below LAQ-8. The means place the later checkpoints close to LAQ-8, with a substantial gap still remaining to full-cache performance. Descriptive wins also depend on the task. Epoch three scores higher than LAQ-8 on nine tasks and lower on seven. Epoch four scores higher on nine, ties on one and scores lower on six.

The largest gain occurs on PassageCount, where both methods have low absolute scores. The largest decline occurs on LCC, whereas RepoBench-P, the other code task, improves by 0.39. Even within a broad category, changing the query construction can favor different examples or tasks. Identical displayed means for epochs three and four likewise conceal changes in individual tasks.

\paragraph{Published baselines at budget 128.}
Table~\ref{tab:main} places our evaluation alongside the Mistral results reported by \citet{liu2025}. Judge Q improves on the other four eviction methods in that experiment. Its score of 42.38 is 3.63 below its full-cache reference. Our corresponding gap is 2.34. The full-cache references already differ by 2.06 points across evaluations. For example, the published full-cache RepoBench-P score is 46.28, compared with 62.56 in our run. Nominal agreement on the backbone and budget therefore leaves appreciable protocol variation. A direct estimate of the gain over Judge Q requires evaluating its checkpoint with the same prompt processing, generation and scoring pipeline.

\section{Why the quality curves can meet}
The probe learns real-valued scores, whereas the decoder receives a set of retained positions. Consider a single head and let $s$ be a reference vector of final selection scores, in decreasing order. Denote the gap at the budget boundary by $\gamma=s_{(B)}-s_{(B+1)}$. For any estimate $\hat{s}$, the sufficient condition
\begin{equation}
       2\|\hat{s}-s\|_\infty < \gamma
       \label{eq:margin}
\end{equation}
preserves the selected set. Indeed, for every retained $i$ and discarded $j$, their estimated difference is at least $\gamma-2\|\hat{s}-s\|_\infty>0$. Consequently two estimators can disagree on attention scores and still pass exactly the same cache to decoding. With identical caches at all heads and layers and identical deterministic decoding, their generated answers coincide.

Equation~\ref{eq:margin} helps interpret the small mean difference between LAQ-8 and FBQ. It gives a mechanism by which further improvements in score estimation can leave answers unchanged. Applying it empirically requires measuring the final selection scores after calibration. A small training MSE averaged over thousands of positions can coexist with a large error on the token ranked 128th. Equal task means can also arise from different answers and opposite gains across tasks, as Table~\ref{tab:main} illustrates. The condition makes no assumption that LAQ achieves the best possible ranking.

\section{Related work}
StreamingLLM retains the initial attention sinks and a recent token window \citep{xiao2024streamingllm}. H$_2$O keeps recent tokens alongside earlier tokens with high accumulated attention \citep{zhang2023h2o}. At prefill, SnapKV uses queries from the end of the prompt to select positions for each head \citep{li2024snapkv}. PyramidKV assigns larger cache budgets to lower layers, where attention is more widely distributed \citep{cai2025pyramidkv}.

LAQ uses pseudo lookahead queries to estimate which cached positions will matter during decoding \citep{wang2025}. Judge Q trains soft tokens to match the prompt attention of actual response tokens \citep{liu2025}. LookaheadKV combines learned lookahead tokens with low-rank adapters to estimate future attention without draft generation \citep{ahn2026}. FBQ studies a narrower parameterization: a prompt-dependent lexical base drawn from the model's own output, an unchanged first query, and seven shared offsets applied at distinct future positions. The zero-offset construction gives a concrete control for learning in this family.

\section{Conclusion}
FTQ-8 turns the first output ID into a block of contextualized future queries. FBQ retains the first query and learns how the remaining seven inputs should deviate from literal repetition. In our Mistral runs, this one-block construction reaches 45.73 against 45.66 for eight-step LAQ. The remaining gap to FullKV and the opposing task-level changes identify where a closer study of the retained evidence is warranted.

\section*{Limitations}
The main comparison covers one model family and one budget. A matched zero-offset run under the main evaluation protocol is needed to isolate the gain due to residual training. The pilot motivates the design but uses a different full-cache reference. The examples \emph{the} and \emph{this} illustrate the motivation. Establishing a specific benefit on generic openings requires stratifying examples by first token and comparing FTQ-8 with FBQ within each group. The reported task aggregates do not provide a paired uncertainty interval for the 0.07-point difference.

Our preliminary system measurements favor FBQ over LAQ in peak memory, time to first token and time per output token. Quantitative claims about the size of these gains require timings and memory measurements for the same checkpoint, hardware, workload and token-emission convention. In particular, a server can emit the committed first token either before or after cache selection, which changes which part of the probe cost appears in first-token latency.

\bibliography{references}
\end{document}